\documentclass[graybox, natbib, nosecnum]{svmult}
\bibpunct{(}{)}{;}{a}{}{,} 

\usepackage{mathptmx}       \usepackage{helvet}         \usepackage{courier}        \usepackage{type1cm}                                    
\usepackage{graphicx}                                     \usepackage{multicol}        \usepackage[bottom]{footmisc}\usepackage[normalem]{ulem}	\usepackage{hyperref}  \usepackage{soul}

\usepackage{times}
\usepackage{helvet}
\usepackage{courier}
\usepackage{url}
\usepackage{graphicx}
\usepackage{dsfont}
\usepackage{xspace}
\usepackage{times}
\usepackage{amsmath,amssymb,amsfonts}
\usepackage{verbatim}
\usepackage{graphicx}
\usepackage{psfrag,graphicx,epsfig,epsf}
\usepackage{graphics}
\usepackage{latexsym}
\usepackage{fancyhdr}
\usepackage{setspace}
\usepackage{color}
\usepackage{url}
\usepackage{mathtools}
\usepackage{amsmath,amsfonts,amssymb,mathrsfs}
\usepackage{psfrag,graphicx,epsfig,epsf}
\usepackage[ruled,linesnumbered, noend]{algorithm2e}
\usepackage{fancyhdr}
\usepackage{wrapfig}
\usepackage{subfigure}
\usepackage{mathtools}
\usepackage{url}
\usepackage{color}
\usepackage{xcolor}
\usepackage{verbatim}
\usepackage{xspace}
\usepackage{placeins}
\usepackage{bm}
\usepackage{caption}
\usepackage[export]{adjustbox}
\usepackage{multirow}
\usepackage{adjustbox}
\usepackage{array}

\newcolumntype{R}[2]{    >{\adjustbox{angle=#1,lap=\width-(#2)}\bgroup}    l    <{\egroup}}

\newcolumntype{P}[1]{>{\centering\arraybackslash}p{#1}}

\def\HiLi{\leavevmode\rlap{\hbox to \hsize{\color{blue!20}\leaders\hrule height .8\baselineskip depth .5ex\hfill}}}

\DeclareMathAlphabet{\mathcal}{OMS}{cmsy}{m}{n}
\newenvironment{aoalgo}[1][H]
{
\begin{algorithm}[#1]}{\end{algorithm}}

\begin{document}

\title*{Asymptotically Optimal\\ Sampling-based Planners}
\author{Kostas E. Bekris and Rahul Shome}
\institute{Kostas E. Bekris \at Rutgers University, New Brunswick, NJ, USA, \email{kostas.bekris@cs.rutgers.edu}
\and Rahul Shome \at Rice University, Houston, TX, USA, \email{rahul.shome@rice.edu}}
\maketitle

\newcommand{\danh}[2][1=]{\todo[linecolor=blue,
			backgroundcolor=blue!5,bordercolor=black,#1]{DH:#2}}
\newcommand{\kb}[2][1=]{\todo[linecolor=green,
			backgroundcolor=green!5,bordercolor=black,#1]{KB:#2}}
\newcommand{\ks}[2][1=]{\todo[linecolor=red,
			backgroundcolor=red!5,bordercolor=black,#1]{KS:#2}}
\newcommand{\rs}[2][1=]{\todo[linecolor=orange,
			backgroundcolor=orange!10,bordercolor=black,#1]{RS:#2}}
\newcommand{\jy}[2][1=]{\todo[linecolor=black,
			backgroundcolor=black!5,bordercolor=black,#1]{JJ:#2}}

\newcommand{\reals}{\mathbb{R}}
\newcommand{\integers}{\mathbb{Z}}

\newcommand{\Wspace}{\mathcal{W}}
\newcommand{\Objects}{\mathcal{O}}
\newcommand{\Manip}{\mathcal{M}}
\newcommand{\nobj}{k}

\newcommand{\Pspace}{\mathcal{P}}
\newcommand{\Pstable}{\mathcal{P}^s}
\newcommand{\pose}{p}
\newcommand{\GeomObj}{\mathcal{WO}}
\newcommand{\Arrange}{\mathcal{A}}
\newcommand{\Pumped}{\mathcal{A^P}}
\newcommand{\pumpedarr}{\alpha^{\mathcal{P}}}

\newcommand{\Qspace}{\mathcal{Q}}
\newcommand{\GeomManip}{\mathcal{WM}}

\newcommand{\Tspace}{\mathbb{T}} 
\newcommand{\Xspace}{\mathbb{X}}
\newcommand{\paths}{\Pi}

\newcommand{\roadmap}{\mathcal{R}}
\newcommand{\graph}{\mathcal{G}}
\newcommand{\nodes}{\mathcal{V}}
\newcommand{\node}{{v}}
\newcommand{\edges}{\mathcal{E}}
\newcommand{\edge}{{e}}
\newcommand{\prmstar}{{\tt PRM$^*$}\xspace}

\newcommand{\rpg}{${\tt RPG}$}

\newcommand{\local}{\mathcal{L}}

\newcommand{\prm}{{\tt PRM}\xspace}
\newcommand{\kprmstar}{{\tt k-PRM$^*$}\xspace}
\newcommand{\rrt}{{\tt RRT}\xspace}
\newcommand{\rrtdrain}{{\tt RRT-Drain}}
\newcommand{\rrg}{{\tt RRG}\xspace}
\newcommand{\krrg}{{\tt k-RRG}\xspace}
\newcommand{\est}{{\tt EST}\xspace}
\newcommand{\rrtstar}{{\tt RRT$^*$}\xspace}
\newcommand{\drrtstar}{{\tt dRRT$^*$}\xspace}
\newcommand{\krrtstar}{{\tt k-RRT$^*$}\xspace}
\newcommand{\srrt}{{\tt RDG}}
\newcommand{\bvp}{{\tt BVP}}
\newcommand{\rdg}{{\tt RDG}}
\newcommand{\lrg}{{\tt LRG}}
\newcommand{\upump}{{\tt UPUMP}}
\newcommand{\prxpump}{{\tt RPG}}
\newcommand{\fixed}{{\tt Fixed}-$\alpha$-\rdg}
\newcommand{\nrob}{k}
\newcommand{\cons}{K}

\newcommand{\frnodes}{V_f}
\newcommand{\frnode}{v_f}
\newcommand{\grnodes}{V_g}
\newcommand{\grnode}{v_g}
\newcommand{\fredges}{E_f}
\newcommand{\fredge}{e_f}
\newcommand{\gredges}{E_g}
\newcommand{\gredge}{e_g}
\newcommand{\kedges}{E_{\cons}}
\newcommand{\kedge}{e_{\cons}}
\newcommand{\safe}{q_s^{\mathcal{M}}}
\newcommand{\hedges}{E_H}
\newcommand{\hedge}{e_H}
\newcommand{\hnodes}{V_H}
\newcommand{\hnode}{v_H}
\newcommand{\hgraph}{H}
\newcommand{\nblank}{b}
\newcommand{\config}{C}
\newcommand{\cquery}{\mathbb{C}}
\newcommand{\pumped}{P}
\newcommand{\pumpedgraph}{\mathcal{G}_P}
\newcommand{\pnodes}{V_P}
\newcommand{\pnode}{v_P}
\newcommand{\pedges}{E_P}
\newcommand{\pedge}{e_P}
\newcommand{\signs}{\Sigma}
\newcommand{\sign}{\sigma}
\newcommand{\gsign}{\sigma_{\pumpedgraph}}
\newcommand{\cedges}{E_c}
\newcommand{\constraints}{\tt c}

\newenvironment{myitem}{\begin{list}{$\bullet$}
{\setlength{\itemsep}{-0pt}
\setlength{\topsep}{0pt}
\setlength{\labelwidth}{0pt}
\setlength{\leftmargin}{10pt}
\setlength{\parsep}{-0pt}
\setlength{\itemsep}{0pt}
\setlength{\partopsep}{0pt}}}{\end{list}}

\newcommand{\dof}{{\tt DoF}}

\newcommand{\mam}{$\mathcal{G}_{\tt MAM}$}
\newcommand{\pr}{\ensuremath{\mathbb{P}}}

\newcommand{\rad}{\ensuremath{r(n)}}
\newcommand{\radstar}{\ensuremath{r^*(n)}}
\newcommand{\radi}{\ensuremath{r_i(n)}}
\newcommand{\radj}{\ensuremath{r_j(n)}}
\newcommand{\crossrad}{\ensuremath{r_R(n)}}
\newcommand{\crossradstar}{\ensuremath{r^*_R(n)}}
\newcommand{\impcrossrad}{\ensuremath{\hat r_R(n)}}
\newcommand{\allimpcrossrad}{\ensuremath{\hat r_{R}(n^R)}}
\newcommand{\ki}{\ensuremath{k_i(n)}}
\newcommand{\kj}{\ensuremath{k_j(n)}}

\newcommand{\mmgraph}{\ensuremath{\mathbb{G}}}
\newcommand{\mmgimp}{\hat\mmgraph}
\newcommand{\mmgexp}{\mmgraph}
\newcommand{\aograph}{\ensuremath{\mathbb{G}^{AO}}}
\newcommand{\tree}{\ensuremath{\mathcal{T}}}
\newcommand{\mmnodes}{\mathbb{\hat V}}
\newcommand{\mmedges}{\mathbb{\hat E}}
\newcommand{\mmnodestpprm}{\mathbb{V}_{\chi_i}}
\newcommand{\mmedgestpprm}{\mathbb{E}_{\chi_i}}
\newcommand{\mmnode}{\mathbb{\hat v}}
\newcommand{\mmedge}{\mathbb{\hat e}}
\newcommand{\sprmstar}{Soft-\ensuremath{ {\tt PRM} }}
\newcommand{\irs}{\ensuremath{ {\tt IRS} }}
\newcommand{\spars}{{\tt SPARS}}
\newcommand{\drrt}{\ensuremath{{\tt dRRT}}}
\newcommand{\dadrrtstar}{\ensuremath{\tt da\_dRRT^*}}

\newcommand{\sig}{{\tt SIG}}
\newcommand{\rmaps}{\ensuremath{\mathfrak{R}}}

\newcommand{\mmprm}{\ensuremath{\text{Random-}{\tt MMP}}}
\newcommand{\astar}{{\ensuremath{\tt A^{\text *}}}}
\newcommand{\mstar}{{\tt M^{\text *}}}
\newcommand{\opens}{P_{Heap}}

\newcommand{\cost}{\mathtt{c}}

\newcommand{\kiril}[1]{{\color{blue} \textbf{Kiril:} #1}}
\newcommand{\chups}[1]{{\color{green} \textbf{Chuples:} #1}}
\newcommand{\rahul}[1]{{\color{red} \textbf{Rahul:} #1}}

\newcommand{\T}{\mathcal{T}}

\newcommand{\leftrm}{\ensuremath{\mathbb{R}_{l}}  }
\newcommand{\rightrm}{\ensuremath{\mathbb{R}_{r}}  }
\newcommand{\leftmetric}{\ensuremath{\mathbb{P}_{l}}  }
\newcommand{\rightmetric}{\ensuremath{\mathbb{P}_{r}}  }
\newcommand{\cfull}{\ensuremath{\cspace_{{\rm full}}}  }
\newcommand{\cfree}{\ensuremath{\cspace_{{\rm free}}}  }
\newcommand{\csmooth}{\ensuremath{\cfree^{smooth}}}
\newcommand{\cobs}{\ensuremath{\cspace_{{\rm obs}}}  }
\newcommand{\cleft}{\ensuremath{\cspace_{{l}}}  }
\newcommand{\cright}{\ensuremath{\cspace_{{r}}}  }
\newcommand{\cshared}{\ensuremath{\cspace_{{s}}}  }
\newcommand{\cgoal}{\ensuremath{q_{{\rm goal}}}  }
\newcommand{\cstart}{\ensuremath{q_{{\rm start}}}  }

\newcommand{\gimpleft}{\ensuremath{\hat\mmgraph_l}}
\newcommand{\gimpright}{\ensuremath{\hat\mmgraph_r}}

\newcommand{\xrand}{\ensuremath{x^{\textup{rand} \ }}}
\newcommand{\xnear}{\ensuremath{x^{\textup{near} \ }}}
\newcommand{\xnew}{\ensuremath{x^{\textup{n}} \ }}
\newcommand{\xlast}{\ensuremath{x^{\textup{last} \ }}}
\newcommand{\xparent}{\ensuremath{x^{\textup{best} \ }}}

\newcommand{\lr}{\ensuremath{\mathbb{R}_{ls}}}
\newcommand{\rr}{\ensuremath{\mathbb{R}_{sr}}}
\newcommand{\lp}{\ensuremath{\mathbb{P}_{l}}}
\newcommand{\rp}{\ensuremath{\mathbb{P}_{r}}}

\newcommand{\motoman}{{\tt Motoman}}
\newcommand{\baxter}{{\tt Baxter}}
\newcommand{\ao}{{\tt AO}\xspace}
\newcommand{\ano}{{\tt AnO}\xspace}
\newcommand{\pc}{{\tt PC}\xspace}

\newcommand\inlineeqno{\stepcounter{equation}\ (\theequation)}

\newcommand{\chomp}{\ensuremath{\tt CHOMP } }

\newtheorem{assumption}{Assumption}

\newcommand{\W}{\mathcal W}
\newcommand\perm[2][\^n]{\prescript{#1\mkern-2.5mu}{}P\_{#2}}
\newcommand\comb[2][\^n]{\prescript{#1\mkern-0.5mu}{}C\_{#2}}
\newcommand{\objectset}{\mathcal{O}}
\newcommand{\object}{o}
\newcommand{\workspace}{\mathcal{W}}
\newcommand{\taskspace}{\mathcal{T}}
\newcommand{\arrangement}{A}
\newcommand{\oar}{p}
\newcommand{\manipulators}{\mathcal{M}}
\newcommand{\manipulator}{\mathit{m}}
\newcommand{\arm}{m}
\newcommand{\arms}{\mathcal{M}}
\newcommand{\taskset}{\mathcal{T}}
\newcommand{\task}{\mathit{T}}
\newcommand{\state}{q}

\newcommand{\Aspace}{\mathcal{A}}
\newcommand{\Afree}{\mathcal{A}_{\rm val}}
\newcommand{\ainit}{A_{\rm init}}
\newcommand{\atarget}{A_{\rm goal}}
\newcommand{\soma}{{\tt soma}}
\newcommand{\coma}{\ensuremath{{\omega}}}
\newcommand{\scoma}{\ensuremath{{{\Omega}}}}
\newcommand{\qset}{\mathcal{Q}}
\newcommand{\startq}{S}
\newcommand{\targetq}{T}

\newcommand{\act}{a}
\newcommand{\actset}{\mathbb{A}}
\newcommand{\moveset}{\bar{\mathcal{O}}}
\newcommand{\home}{Q}
\newcommand{\scomaset}{\{\scoma\}}
\newcommand{\tour}{{\Gamma}}
\newcommand{\tspgraph}{\graph_{\tour}}
\newcommand{\tspnodes}{\nodes_{\tour}}
\newcommand{\tspedges}{\edges_{\tour}}

\newcommand{\kuka}{{\tt{Kuka }}}
\newcommand{\sininv}{\sin^{-1}}
\newcommand{\cosinv}{\cos^{-1}}
\newcommand{\milp}{{\tt{MILP}}\xspace}
\newcounter{model}
\newenvironment{model}
{\refstepcounter{model}}
{\begin{center}
\textbf{Model. }~\themodel
\end{center}
\medskip}
\definecolor{darkgreen}{RGB}{30,150,30}
\newcommand{\commentdel}[1]{{ #1}}
\newcommand{\commentadd}[1]{{\color{darkgreen} #1}}

\newcommand\blfootnote[1]{  \begingroup
  \renewcommand\thefootnote{}\footnote{#1}  \addtocounter{footnote}{-1}  \endgroup
}
\newcommand{\cspace}{\mathcal{Q}}
\newcommand{\carms}{\cspace_{\arms}}
\newcommand{\objectmotion}{\mathbb{A}}
\newcommand{\orbit}{\mathbb{O}}
\newcommand{\deltaint}{\cspace_{\delta}}
\newcommand{\closedeltaint}{\overline{\deltaint}}
\newcommand{\closedfree}{\overline{\cfree}}

\newcommand{\Chi}{\mathcal{X}}
\newcommand{\perc}{\Big(\frac{\log n}{n}\Big)^{\frac{1}{d}}}
\newcommand{\ball}[2]{{\mathcal{B}}_{#2}({#1})}
\newcommand{\cone}[4]{{\mathcal{H}}_{#3}({#1},{#2},{#4})}
\newcommand{\neigh}[2]{{\mathcal{N}}_i({\mathcal{B}}_{#2}({#1}))}
\newcommand{\fail}[2]{{\mathcal{F}}_{n}({\mathcal{B}}_{#2}({#1}))}
\newcommand{\nearR}{\mathcal{R}_n}
\newcommand{\trans}{t}
\newcommand{\settrans}{\mathcal{T}}
\newcommand{\mode}{\mathcal{{M}}}
\newcommand{\alg}{\mathtt{ALGO}}
\newcommand{\qstart}{\state_{0}}
\newcommand{\qtarget}{\state_{1}}
\newcommand{\rgg}{\widehat{\mathcal{G}}}
\newcommand{\rgv}{\widehat{\mathcal{V}}}
\newcommand{\rge}{\widehat{\mathcal{E}}}
\newcommand*{\QEDA}{\hfill\ensuremath{\square}}\newcommand{\disp}{\mathcal{D}_n^q}
\newcommand{\dispstart}{\mathcal{D}_n^{\qstart}}
\newcommand{\dispval}{\mathcal{R}_n}
\newcommand{\oneball}{\mu_1}
\newcommand{\mufree}{\mu_{free}}
\newcommand{\dispersion}{\mathtt{disp}_n}
\newcommand{\mudelta}{\mu_\delta}
\newcommand{\deltan}{\delta_n}
\newtheorem{construction}{Construction}

\newcommand{\rpn}{r^{\prime}_n}

\newcommand{\traj}{\sigma}
\newcommand{\trajobs}{\overline{\sigma}}
\newcommand{\trajset}{\Sigma}
\newcommand{\realnum}{\mathbb{R}}
\newcommand{\initstate}{\state_{\mathrm {init}}}
\newcommand{\goalset}{\cspace_{\mathrm {goal}}}
\newcommand{\goalstate}{\state_{\mathrm {goal}}}
\newcommand{\trajopt}{\traj^*}
\newcommand{\algo}{{\tt{ALG}}\xspace}
\newcommand{\knear}{{\tt k}-near\xspace}
\newcommand{\costrand}{\mathfrak{C}}

\definecolor{darkblue}{RGB}{47,82,143} 
\section{Synonyms}
\textit{AO planning, optimal motion planning}

\section{Definition}

An asymptotically optimal sampling-based planner employs sampling to
solve robot motion planning problems and returns paths with a cost
that converges to the optimal solution cost, as the number of samples
approaches infinity.

\section{Overview}
Sampling-based planners sample feasible robot configurations and connect them with valid paths. They are widely popular due to their simplicity, generality, and elegance in terms of analysis. They scale well to high-dimensional problems and rely only on well-understood primitives, such as collision checking and nearest neighbor data-structures. Roadmap planners, e.g., the \textit{Probabilistic Roadmap
Method} (\prm) \citep{kavraki1994probabilistic}, construct a graph, 
where nodes are configurations and edges are local paths. The roadmap can be
preprocessed and used to answer multiple queries. Alternatives, e.g., the \textit{Rapidly Exploring Random Tree} (\rrt) \citep{lavalle2001randomized}, build a
tree and aim to quickly explore the reachable configurations for solving a specific query.  Tree-based variants can deal with challenges that involve significant dynamics, where there may not be access to a method (i.e., a steering function) for connecting two configurations with a local path.

Many early methods aimed to provide \textit{probabilistic completeness} \citep{hsu1999path}, i.e., a solution will be found, if one exists, as the number of sampled configurations approaches infinity. Further analysis focused on properties of configuration spaces, which allowed sampling to be effective, such as \textit{$ \epsilon
$-goodness} \citep{kavraki1998randomized} and \textit{expansiveness} \citep{hsu1999path}. A large number of variants were proposed for enhancing the practical performance of sampling-based planning \citep{amato1998obprm, kim2003extracting, raveh2011little}.  These early methods did not focus on returning paths of high-quality. As the planners became increasingly faster, however, the focus transitioned towards understanding the conditions under sampling-based planners can asymptotically converge to paths that optimize a desirable cost function. This gave rise to a new family of sampling-based planners, which can also achieve \textit{asymptotic optimality}.

Consider a robot in a workspace with obstacles, where a configuration $ \state $ is a vector of robot variables that define the workspace volume occupied by the robot.

\vspace{-.05in}
\begin{definition}[Configuration Space]
All possible configurations of the robot define a set
$ \cspace \subset \realnum^d $. The feasible subset
$ \cfree \subseteq \cspace $ and the infeasible subset
$ \cobs \subseteq \cspace $ refer to configurations that do not result
or cause collisions with obstacles in the workspace, respectively.
\end{definition}
\vspace{-.05in}

The notion of collisions can be generalized to express any feasibility
constraint. 

\vspace{-.05in}
\begin{definition}[Feasible Paths with Strong $\delta$-clearance]
	\label{def:feasible}
A feasible path $ \traj : [0,1] \rightarrow \cfree $ is a
parameterized continuous curve of bounded variation. The set of all
possible such paths is $ \trajset $. A feasible path $\traj$ has
strong $\delta$-clearance, if $\traj$ lies entirely inside the $\delta$
interior of $\cfree$, i.e., $\forall \tau \in [0,1]: \traj(\tau) \in
int_{\delta}(\cfree)$.
\end{definition}
\vspace{-.05in}

The \textit{$ \delta $-clearance} property guarantees there is always
a $ d $-dimensional $ \delta $-ball in $\cfree$ - and, thus, a
positive volume - around configurations of a solution path.

\vspace{-.05in}
\begin{definition}[Robustly Feasible Motion Planning]
Given $ \initstate \subseteq \cfree$ and a set
$ \goalset \subseteq \cfree$, the robustly feasible motion planning
problem $ (\cfree, \initstate, \goalset) $ asks for a feasible path
$ \traj $ with strong $\delta$-clearance, so that $ \traj[0]
= \initstate$ and $\traj[1] \in \goalset $.
\end{definition}
\vspace{-.05in}

\begin{definition}[Optimal Motion Planning]
Given a path cost function $ \cost
: \trajset \rightarrow \realnum_{\geq 0} $, an optimal solution to the
motion planning problem satisfies: $ \trajopt
= \underset{\sigma}{\mathtt{argmin}} \ \cost(\sigma) $.
\end{definition}
\vspace{-.05in}

The optimal solution need not be unique but the minimum cost is unique
and finite. Let $\costrand_n^{ALG}$ define the extended random variable
corresponding to the cost of the minimum-cost solution returned by
algorithm $ALG$ after $n$ iterations.

\begin{figure}[h]
	\centering 
	\includegraphics[width=0.9\textwidth]{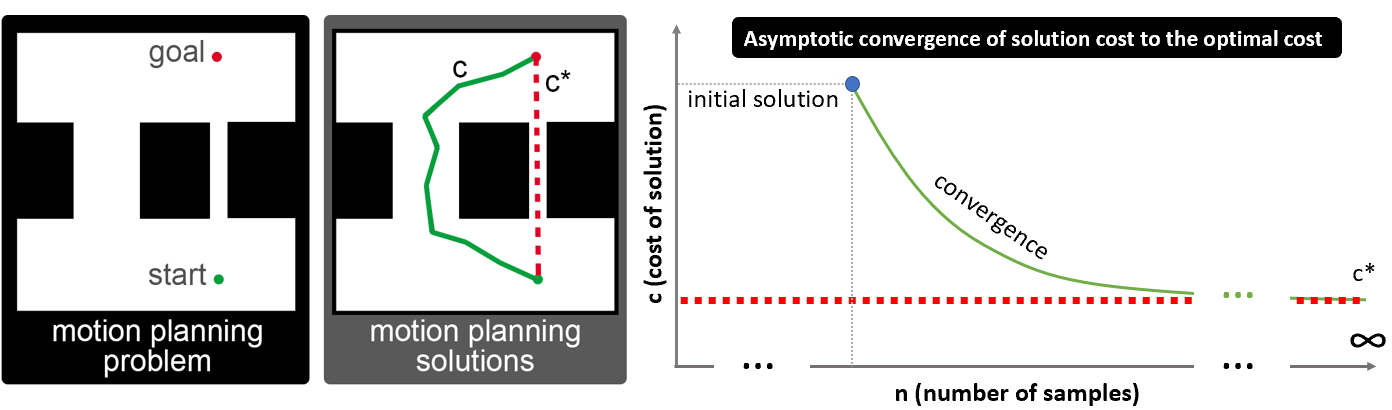}	
	\vspace{-0.1in}
	\caption{\small 
	A motion planning problem \textit{(left)} consists of the \textit{start}, and \textit{goal} in $\cfree$ (shown in white), while $\cobs$ obstacles are black. A solution to the problem \textit{illustrated in the middle} has a cost $c$ (green) relative to the optimal solution cost $c^*$ (red dashed). The plot on the right illustrates the asymptotic convergence of the solution cost $\cost$ to the optimal one $\cost^*$ over the number of samples $n$. 
	}
\vspace{-0.2in}
\label{fig:conv}
\end{figure}

\vspace{-.05in}
\begin{definition}[Asymptotic Optimality]
An algorithm is asymptotically optimal if, for a robustly feasible
motion planning problem $ (\cfree, \initstate, \goalset) $, which
admits a robustly optimal solution with finite cost $c^* = \cost(\trajopt)$ 
ensures that:
\vspace{-.1in}
$$P(\{\limsup_{n \rightarrow \infty}\ \costrand_n^{ALG} = c^*\}) = 1.$$
\end{definition}
\vspace{-0.1in}
The corresponding literature describes the guarantee of asymptotic optimality
in terms of this event occurring asymptotically. This is highlighted in Fig \ref{fig:conv}.
\vspace{-0.3in}

\begin{comment}

describe the necessity of every point in $ \cfree $ to always be
continuously connected to a positive volume of $ \cfree $. The measure
of these volumes of $ \cfree $ affect the probability of sampling
inside them. Such well-behaved spaces allow the sampling process to
provide a positive probability of exploring parts of the space
necessary for solution discovery.

The notion of \textit{probabilistic completeness} focuses on
finding \textit{any} feasible solution. While an important necessary
step in the evolution of motion planning algorithms, this property
ignored any quality guarantee for the returned solutions. In a vast
array of practical applications, it does not suffice to only find a
solution, and the high-quality solutions are not only desirable but
also necessary.

also \textit{probabilistically complete}(PC), but theoretically an
important distinction between PC algorithms and AO planners is
that, \textit{asymptotic optimality} cannot only contend with
reasoning about reaching the desired goal but additionally requires a
characterization of the desired optimal path $ \trajopt $ which
minimizes the cost $ \cost(\trajopt) $ of such a path.

that can be controlled by affecting changes to its state in the world,
where the subset of the world wherein the robot coexists with static
geometries called \textit{obstacles} is called a workspace, which
typically can be described as $ \workspace\subset SE(3) $.
\end{comment} 

\section{Key Research Findings}

Table~\ref{tab:stateoftheart} summarizes different arguments for \ao motion planning. Broadly, problems can be separated into kinematic ones where pairs of samples can be connected, and kinodynamic problems with non-trivial dynamics that do not allow such connections. Kinematic algorithms were proposed first, e.g., \prmstar, \rrtstar ~\citep{karaman2011sampling}, {{\tt FMT}$^*$~\citep{janson2015fast} and then \ao kinodynamic methods were introduced, e.g., {\tt SST}$^*$~\citep{li2016asymptotically}, {\tt AO-}$\mathcal{A}$~\citep{hauser2016asymptotically, kleinbort2019rrt2}. The analyses also differ regarding the nature of the convergence property. Critical data-structures for these algorithms are also included in the table.

\begin{table}[]
\begin{tabular}{c|p{0.18\textwidth}|p{0.16\textwidth}|p{0.12\textwidth}|p{0.44\textwidth}}
              & \textbf{Convergence} & \textbf{Structure} & \textbf{Algo} & \textbf{Condition} \\ \hline
\multirow{4}{*}{\rotatebox[origin=c]{90}{Kinematic$\quad\quad\quad\quad$}}   & Almost sure          & Roadmap                 & PRM*          & {\small $ r_n \ >\ 2(1+\frac{1}{d})^{\frac{1}{d}}  \Big( \frac{\mu(\cfree) }{\zeta_d}\Big)^{\frac{1}{d}} \perc  $   \newline     $ k_n \ >\ e(1+\frac{1}{d})\log n  $
}  \\ \cline{2-5} 
                             & In probability       & Search tree on Roadmap                 & FMT*          & {\small $ r_n \ >\ 2(\frac{1}{d})^{\frac{1}{d}}  \Big( \frac{\mu(\cfree) }{\zeta_d}\Big)^{\frac{1}{d}} \perc  $   }                  \\ \cline{2-5} 
                             & In probability       & Tree  with rewiring                  & RRT*          & {\small $ r_n \ \geq\ (2+\theta)\Big( \frac{(1+\frac{\epsilon}{4})c^*}{(d+1)\theta(1-\nu)} \Big)^{\frac{1}{d+1}}$\newline$ \Big( \frac{\mu(\cfree) }{\zeta_d}\Big)^{\frac{1}{d+1}} \Big( \frac{\log n}{n} \Big)^{\frac{1}{d+1}}  $  }                 \\ \cline{2-5} 
                             & Deterministic, dispersion-based       & Roadmap                 & PRM*, FMT*    & If sampling dispersion is {\small $\mathcal{O}\Big(\Big(\frac{1}{n}\Big)^{\frac{1}{d}}\Big)$}, then {\small$r_n \in \omega\Big(\Big(\frac{1}{n}\Big)^{\frac{1}{d}}\Big)$}                 \\ \hline
\multirow{3}{*}{\rotatebox[origin=c]{90}{Kinodynamic$\quad\quad$}} & In probability       & Forward search tree                    & SST*          & Random selection, Monte Carlo Prop: random control and duration        \\ \cline{2-5} 
                             & In probability       & Forward search tree                    & AO-RRT        & RRT selection in augmented state-cost space, Monte Carlo propagation           \\ \cline{2-5} 
                             & In probability       & Meta-algo               & AO-{$\mathcal{A}$}          & Repeatedly call a PC algorithm {$\mathcal{A}$} with lowering cost-bound                 \\ \hline
\end{tabular}
\vspace{-.1in}
\caption{\small Summary of state-of-the-art results on asymptotic optimality. $r_n$ and $k_n$ describe the connection neighborhood for roadmaps as a radial region and number of nearest neighbors. $n$ is the number of samples, $d$ is the dimensionality of $\cfree$, $\mu$ is the volumetric measure, $\zeta_d$ is the volume of an unit $d$-dim. ball, $\epsilon \in (0,1)$ is the error from the optimum. For \rrtstar: $\theta\in(0,\frac{1}{4}), \nu\in(0,1)$. Kinodynamic planners do not require geometric neighborhood definitions. 
}
\vspace{-.3in}
\label{tab:stateoftheart}
\end{table}

\vspace{0.1in}

\renewcommand{\arraystretch}{1.5}

For all methods employing random sampling, $\cost_n$ is a random variable that depends on the realization of $n$ i.i.d samples. Consider an arbitrarily small error measure $\epsilon>0$. Then, failing to converge corresponds to $\cost_n > (1+\epsilon)\cost^*$. In terms of the type of convergence, "almost sure convergence" dictates that out of all the realizations of the algorithm as $n$ reaches infinity, the failure event $\{ \cost_n > (1+\epsilon)\cost^* \}$ is assured to occur \textit{a finite number of times}, i.e., there exists a large enough $n_0\gg1$ such that $\forall n>n_0,\ \cost_n \leq (1+\epsilon)\cost^*$. "Convergence in probability" only requires that at infinity the probability of the failure event goes to zero. Deterministic convergence can be argued geometrically given guarantees for the dispersion of samples. This means that the error is surely upper bounded by $\epsilon$ for a large enough $n$. These convergence results hold for all (arbitrarily small) values of $\epsilon$.

\vspace{0.1in}
\noindent\textbf{Analysis Model: Random Geometric Graphs}

\begin{wrapfigure}{r}{0.52\textwidth}
	\vspace{-0.3in}
	
	\begin{aoalgo}[H]
		\label{algo:prm}
		\caption{\textbf{PRM$^*$ ($ \initstate,\goalset,n $)}}
		$ \graph(\nodes =  \{ \initstate\cup \goalset \}, \edges = \emptyset) $\;
		\lFor{$ n\ times $}
		{
			$ \nodes \gets \nodes \cup{\mathtt{sample}()} $
		}
		\For{$ \node \in \nodes $}
		{
			\For{$ u \in {\mathtt{Neighborhood}(\node, \graph)} $}
			{
				\If{$ \mathtt{valid\_edge}(u,\node) $}
				{
					$ \edges\gets\edges\cup\{(\node,u),(u,\node)\} $\;	
				}
			}
		}
		$ \mathbf{return}\ {\mathtt{A^*}}(\initstate,\goalset,\graph) $
	\end{aoalgo}
	
	\begin{aoalgo}[H]
		\label{algo:rrt}
		\caption{\textbf{RRT$^*$ ($ \initstate, \goalset, n $)}}
		$ \tree(\nodes=\{\initstate\}, \edges=\emptyset) $\;
		\For{$ n \ times$}
		{
			$ \node_{\mathrm{rand}} \gets \mathtt{sample}() $\;
						$ u_{\mathrm{near}} \gets \mathtt{near}(\node_{\mathrm{rand}}, \tree)$\;
			$ \node \leftarrow \mathtt{steer}(u_{\mathrm{near}}, \node_{\mathrm{rand}}) $\;
						$ \mathcal{N} \gets \emptyset $\;
			\For{$ u \in {\mathtt{Neighborhood}(\node, \tree)} $}
			{
				\If{$ \mathtt{valid\_edge}(u,\node) $}
				{
					$ \mathcal{N}\gets\mathcal{N}\cup \{u\} $\;	
				}
			}
			$ u_{\mathrm{best} } \gets  \underset{ u\in \mathcal{N} }{argmin}\  \mathtt{C}(u,\tree)+\cost(\overrightarrow{u \node}) $\;
			$ \nodes \gets \nodes \cup \{\node\}\ ;\ \edges \gets \edges \cup \{(u_{\mathrm{best}}, \node)\}$\;
																											$ \tree\gets \mathtt{rewire}(\node,\mathcal{N},\tree) $\;
		}
		$ \mathbf{return}\ {\mathtt{trace}}(\goalset,\tree) $\;
	\end{aoalgo}
	
	\vspace{-0.35in}
\end{wrapfigure}

\noindent Identifying how 
sampling-based motion planners can
achieve \textit{asymptotic optimality} was first
achieved by building on top of results in
random geometric graphs 
. It resulted in new algorithms that provide this property, such as \prmstar\
and \rrtstar \citep{karaman2011sampling}.

\noindent\textit{Outline of\ \prmstar and \rrtstar}: Algo~\ref{algo:prm} describes \prmstar: it resembles \prm except for the functional description of the neighborhood $\mathcal{N}$ in terms of $r_n$ or $k_n$ (Table~\ref{tab:stateoftheart}). Algo~\ref{algo:rrt} explains \rrtstar: the difference with \rrt is a rewiring step that reasons about local connections between vertices within a neighborhood $\mathcal{N}$ given a radius $r_n$  (Table~\ref{tab:stateoftheart}). Figure~\ref{fig:algos} shows that this radius heavily affects the solutions returned.

\begin{figure}[h]
    \vspace{-0.1in}
	\centering 
	\includegraphics[width=0.95\textwidth]{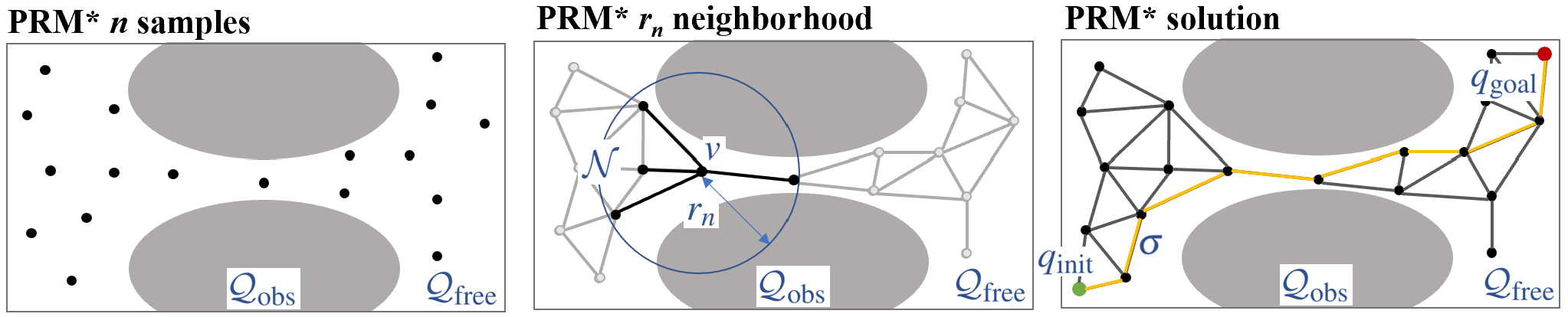}	
	\includegraphics[width=0.95\textwidth]{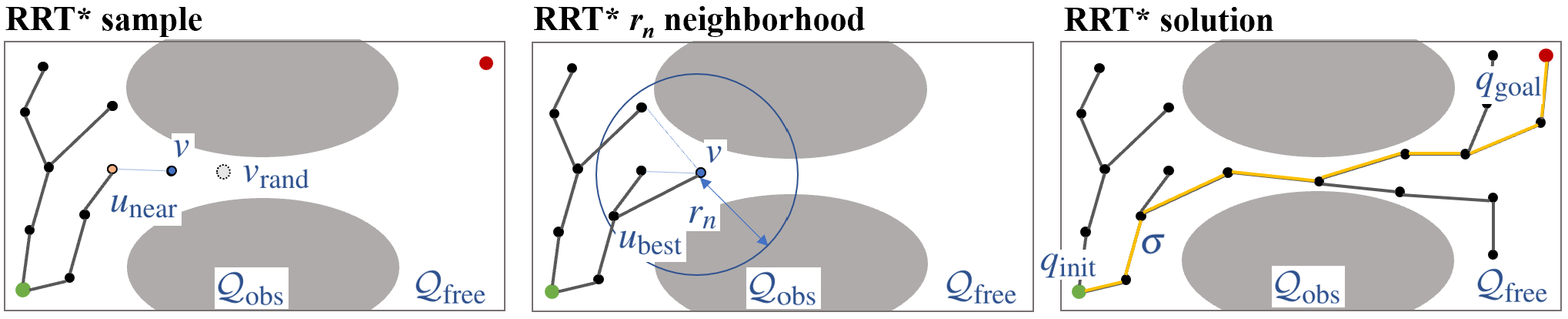}	
	\caption{\small \textbf{\textit{(Top)}}: Different steps of \prmstar: \textit{(Left)} the $ n $ random samples; \textit{(Middle)} The $ r_n $ neighborhood of each sample \node\ is checked for a connection; \textit{(Right)} The solution is \textit{traced} over $ \graph $.\\		
	\textbf{\textit{(Bottom)}}: Different steps of \rrtstar: \textit{(Left)} Dashed circles represent the random sample, the nearest node of the tree, and the result (blue point \node) of \textit{steering} towards the random sample; \textit{(Middle)} The $ r_n $ neighborhood of \node\ is checked for the best connection; \textit{(Right)} The solution is \textit{traced} over $ \tree $.
	}
	\label{fig:algos}
	\vspace{-.3in}
\end{figure}

These planners operate over underlying random geometric graphs. The graph theory literature describes properties of such graphs \citep{penrose2003random}, including  connectivity and percolation. $ \graph_n^r $ is surely connected when
{\small $ r > \Big(\frac{1}{\zeta_d}\Big)^{\frac{1}{d}}{\Big(\frac{\log n}{n}\Big)}^{\frac{1}{d}}$}. This threshold relates to AO requirements in motion planning as long as the graph is connected in the vicinity of the optimal path $ \trajopt $.

\vspace{-.05in}
\begin{definition}[Random Geometric Graph for Motion Planning]
The vertices of a random geometric graph $ \graph_n^r(\nodes, \edges)
$ correspond to $ n $ i.i.d. uniformly sampled configurations in
$ \cfree $. Each vertex is connected with edges that lie in $ \cfree $
to all configurations that are within a distance $r \in \realnum_{>
0}$ away.
\end{definition}
\vspace{-.05in}

	\begin{wrapfigure}{r}{0.45\textwidth}
	\centering 
	\includegraphics[width=0.44\textwidth]{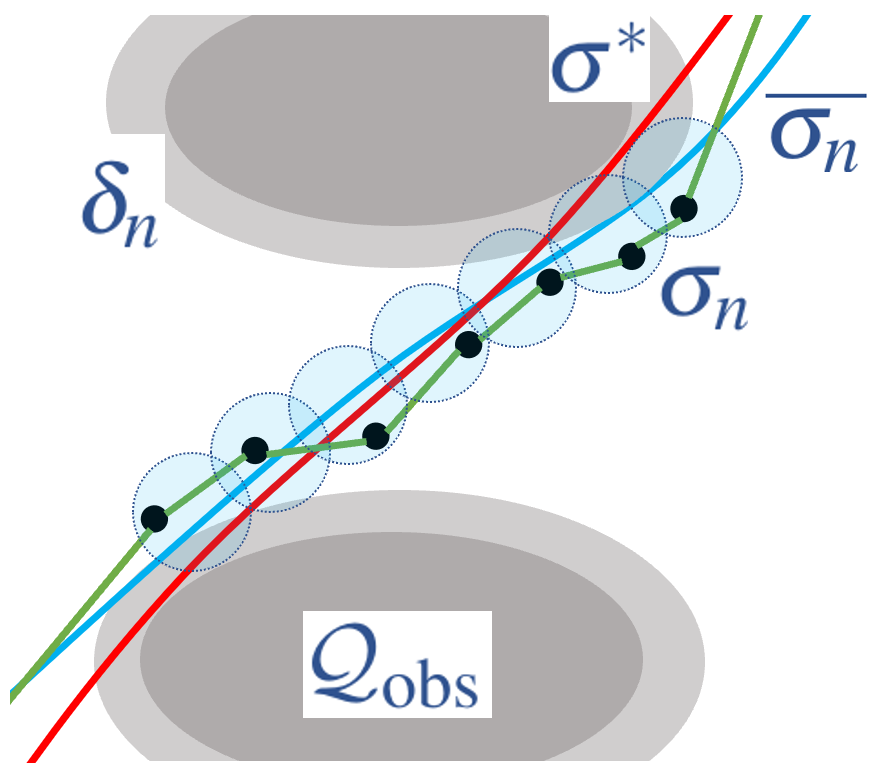}	
	\vspace{-.1in}
	\caption{\small The optimal path $ \trajopt $ in red can touch the boundary of $ \cobs $. The clearance $ \delta_n $ (light gray around the obstacles), and an  \textit{observing path} $ \trajobs_n $ (in blue) can be studied by \textit{tiling} $ \trajobs_n $, i.e., defining a sequence of overlapping \textit{hyperballs}. A solution $ \traj_n $ (in green) connects consecutive \textit{tiles}.
	}
	\label{fig:construction}
	\vspace{-0.3in}
\end{wrapfigure}

\noindent \textbf{Outline of the Analysis: } Given the \textit{$ \delta $-clearance} of any feasible solution $ \traj_n $ (Definition~\ref{def:feasible}), there exists a volume of $ \cfree $ surrounding $ \traj_n $. Typically, the optimal path $ \trajopt
$ itself can touch the $ \cfree $ boundary and hence
possesses \textit{0-clearance} at such contact points. As long
as there exists a sequence of $ \trajobs_n $ paths, each having
 \textit{$ \delta_n $-clearance} for $ \delta_n > 0$, such that
$ \lim\limits_{n\rightarrow\infty} \cost(\trajobs_n)
= \cost(\trajopt)$, the algorithm can operate over a positive volume
around each $ \trajobs_n $. This positive volume provides positive probability of sampling in these regions. The sequence of $ \{ \trajobs_n \} $ is called a sequence
of \textit{observing paths}. As $ n $ increases,
$ \delta_n $, and $ r_n $ decrease, meaning that $ \delta_n $
expresses the algorithm's ability to discover solutions
through increasingly narrower \textit{corridors} of $ \cfree $.

Each $ \trajobs $ has a volume of $ \cfree $ given its
clearance $ \delta_n $. The volume surrounding an observing
path $ \trajobs $ is divided into a \textit{finite} sequence
of \textit{tiling} constructions (hyperballs \cite{karaman2011sampling,janson2015fast}, or hypercubes \cite{solovey2018new}), where each \textit{tile} has positive
volume. By setting an appropriate $ r_n $ value, the algorithm ensures connectivity in the $ \delta_n $-clearance volume of each
observing path $ \trajobs_n $. This defines a solution path $ \traj_n $
along consecutively connected configurations along $ \trajobs_n $. The \knear analysis broadly operates over the expected number of samples in
the volumes described by the $ r_n $ variants.

The probability of discovering a $\traj_n$ close enough to a $\trajobs_n$, which in turn can get arbitrarily close to some desired $\trajopt$ can be described in terms of the  probability of sampling along the construction. At this stage, the functional estimate of $r_n$ depends on the nature of asymptotic convergence desired. This explains the difference between $r_n$ described by \prmstar for almost sure convergence, versus {\tt FMT}$^*$~\citep{janson2015fast} described for convergence in probability. The latter analysis~\citep{janson2015fast} also deduced the convergence rate bound for \prmstar and {\tt FMT}$^*$ as $\mathcal{O}(n^{-\frac{1}{d+\rho}})$, when the algorithm is executed for $n$ samples in a $d$-dim. configuration space, where $\rho$ is an arbitrarily small constant. The convergence rate of \ao sampling-based algorithms is an important factor, which dominates the finite time properties and practical performance of solutions.

Recent work  \citep{solovey2019revisiting} argues that tree-based \ao sampling-based motion planners need to account for the existence of a chain of samples from the start along the hyperball tiling, for every hyperball, in addition to ensuring that the hyperball has a sample in it with increasing $n$. In a way, \textit{time} is an additional dimension to deal with. This leads to an $r_n$ where the exponent changes to $\frac{1}{d+1}$ for \rrtstar (Table~\ref{tab:stateoftheart}). 

\vspace{0.1in}
\noindent\textbf{Analysis Model: Batched and Deterministic Sampling}

\begin{wrapfigure}{r}{0.45\textwidth}
\vspace{-.3in}
	\includegraphics[width=0.44\textwidth]{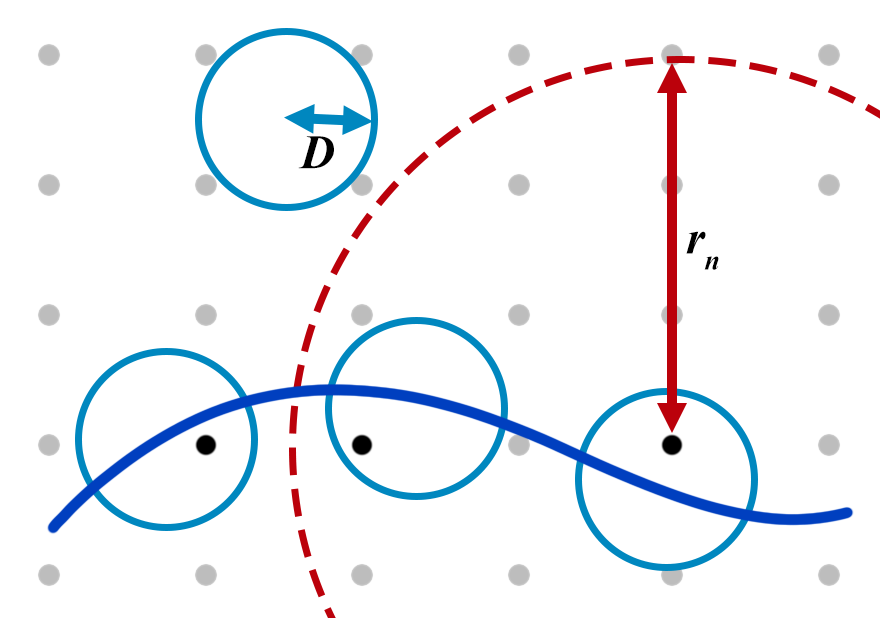}
	\vspace{-.1in}
    \caption{\small A deterministic sampling sequence with dispersion $D$, and tiling fo the path. The radius $r_n$ can be described in terms of $D$.}
	\vspace{-0.3in}
\end{wrapfigure}

Low-dispersion deterministic sampling sequences guarantee samples with dispersion  \citep{janson2018deterministic} $ D(\nodes) = \mathcal{O}(n^{\frac{1}{d}})$, where $\nodes$ is the set of nodes of the planner. Dispersion is defined as the radius of the largest empty hyperball, which does not contain a sample. This is tighter than the expected dispersion from uniform sampling.  As already mentioned, asymptotic guarantees are closely related to the ability to successfully sample within a sequence of tiling hyperballs along a path. With uniform sampling, the success of this event is probabilistic, and depends on the volume of the hyperballs in the sampling domain. If the dispersion of the determinstic sequence is guaranteed to be lower than the radius of the hyperball, a sample \textit{is assured} inside the hyperball. Once samples are assured, the connection radius needs to only maintain connections between consecutive hyperballs. Results provide that an algorithm is \ao if $\underset{{n\rightarrow\infty}}{\lim}\ D(\node)\cdot r_n \rightarrow \infty $. The same line of work also expressed the connection radius bounds necessary for acceptable suboptimality error bounds, since such an error can express hyperball regions that admit such low-error solutions. It also provides a set of relations between dispersion and convergence rates as well as computational complexity results, including tighter bounds with $\epsilon$-net sampling~\citep{tsao2019sample}.

\vspace{0.1in}
\noindent\textbf{Analysis Model: Monte Carlo Trees}

A line of work focused on systems with dynamics and removed the requirement for a steering function by starting from first principles to achieve \ao properties \citep{li2016asymptotically,littlefield2018efficient}. 
The analysis starts with a Monte Carlo search tree, which performs \textit{random selection} of nodes and \textit{random propagation} of controls and does not depend on a steering function. Such an approach is shown to be \ao, but also not practical due to slow convergence.  A {\tt Best-Near} variant of the Monte Carlo tree, however, was shown to be both asymptotically (near-)optimal and have practical convergence. The variant prioritizes the selection of nodes with good path quality within a neighborhood of the random sample and still uses Monte Carlo propagation. It can achieve \ao\ properties by reducing the neighborhood size for this selection choice over computation time. The variant can be further improved in practice by sparsifying the underlying tree data structure and storing only nodes that locally have good path quality, giving rise to the {\tt SST} algorithm (Fig ~\ref{fig:sst}). Heuristics can further speed up practical performance \citep{littlefield2018efficient}.

\begin{figure}[h]\vspace{-0.25in}
    \centering
            \includegraphics[height=1.5in]{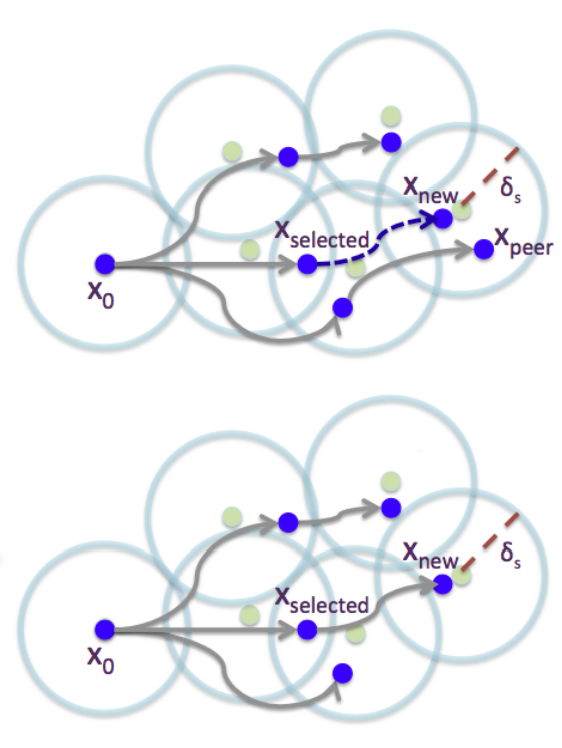}
        \includegraphics[height=1.5in]{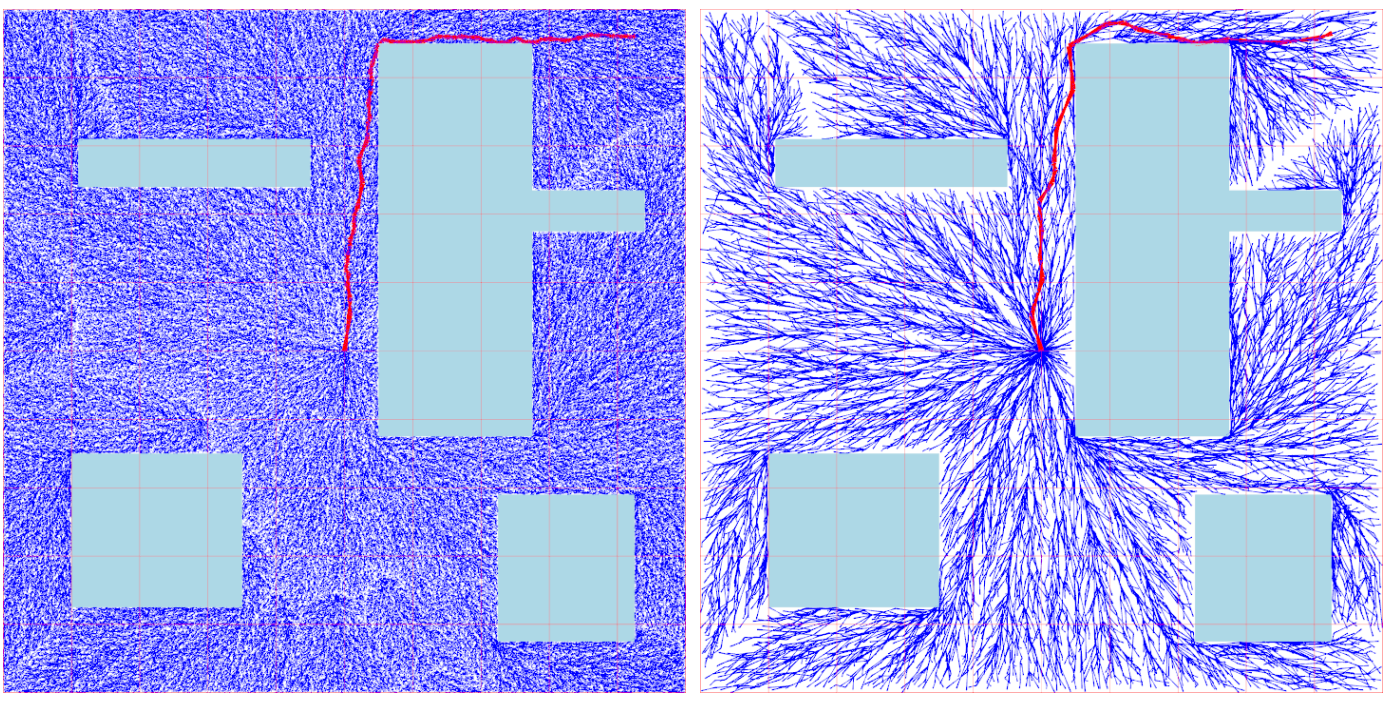}
    \vspace{-.1in}
    \caption{\small \textit{(Left):} The {\tt SST} approach propagates the  node with the best path cost from the start in a neighborhood of a random sample to achieve asymp. near-optimality. For computational efficiency purposes, it also prunes nodes that are locally dominated in terms of path quality. (\textit{Right}): A dense tree constructed using \rrtstar (Algo \ref{algo:rrt}) and then the sparser tree from {\tt SST} \citep{li2016asymptotically}.}
    \label{fig:sst}
\vspace{-0.25in}
\end{figure}

The argument reasons about hyperballs along robust feasible paths, which are defined for some arbitrarily small $\epsilon$. The search tree has to discover a branch connecting consecutive hyperballs along the feasible path. Given an assumed smootheness in the dynamics, random controls and durations are sufficient to connect consecutive regions with probability measures independent of $n$. Then, it is sufficient to show that the algorithm can select every node infinitely often to allow opportunities to sample the desired edge using such Monte Carlo propagation. This process continues till the goal region is reached asymptotically.

\vspace{0.1in}
\noindent\textbf{Analysis Model: Search in State-Cost Space} 

\begin{wrapfigure}{r}{0.41\textwidth}
\vspace{-.4in}
	\includegraphics[width=0.4\textwidth]{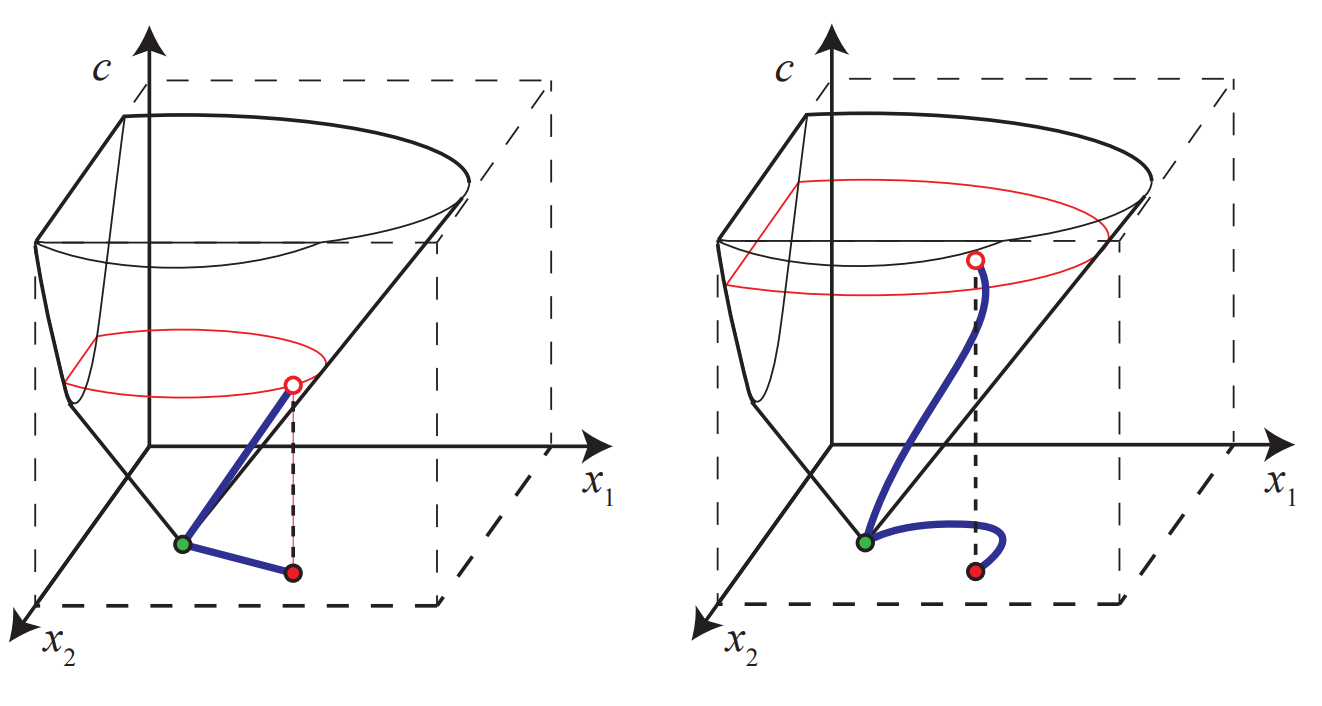}
	\vspace{-.1in}
    \caption{\small Two solutions in 2D $\{x_1,x_2\}$ but with different costs, showing up as the difference in the state-cost space $\{ x_1,x_2,c \}$ \citep{hauser2016asymptotically}.}
	\vspace{-0.3in}
\end{wrapfigure}

\noindent Another direction for asymptotic optimality is a meta-algorithm, referred to as \textbf{$ \ao-\mathcal{A} $} \citep{hauser2016asymptotically}.  The approach provides \ao properties as long as a \textit{probabilistically complete} (\pc) algorithm $ \mathcal{A} $, which achieves \textit{exponential convergence}, is applied in the \textit{state-cost space}. An algorithm is \pc in the \textit{state-cost space} if it is guaranteed to find a solution within a desired cost-bound, if one exists.  The $ \ao-\mathcal{A} $ algorithm inspects the \textit{state-cost} space, and repeatedly calls  algorithm $ \mathcal{A} $ to produce a solution within the best cost bound discovered so far. Algorithm $\mathcal{A}$ takes the cost-bound as an argument and returns the first solution discovered within the input cost bound. By reducing the cost bound, the solution converges to the optimal cost. The \pc and exponential convergence properties of $ \mathcal{A} $ ensure that the expected running time is finite.

A kinodynamic {\tt AO-RRT} ~\citep{kleinbort2019rrt2}, which uses the virtues of randomness laid out in Monte-Carlo trees, but operates directly in the state-cost space for obtaining a solution with a single invocation of the algorithm was also studied. The selection procedure is the same as \rrt in the state-cost space. The cost is instrumental is growing parts of the tree within the sampled cost bound. This ensures that as the cost of the best solution keeps decreasing, only the part of the search tree that can improve the solution is given a chance to grow using Monte-Carlo propagation. The argument formulates the probabilities of the tree traversing a construction of hyperballs as a \textit{Markov chain}. The probability measures need to be shown to be independent of $n$. This simplified argument can then deduce that the random process reaches the sink-node of the chain asymptotically, i.e., it reaches the goal region following the sequence of regions along the construction.

\vspace{0.1in}
\noindent\textbf{Bridging Theoretical Guarantees and Practical
  Performance}

\textit{Asymptotic optimality} comes at the cost of computational overhead per iteration when compared to \textit{probabilistically complete} or heuristic alternatives, which motivated work on balancing this desired property with practical performance.

\textbf{Relaxed but Practical Guarantees:} Computational trade-offs can be made by foregoing \ao properties for \textit{asymptotic
near-optimality}\citep{marble2013asymptotically}. Such tradeoffs can utilize ideas from \textit{graph spanners}, which refer to subgraphs that trade-off connectivity with bounded suboptimality. This trade-off is defined by a parameter that bounds the acceptable path cost degradation relative to the inclusion of edges during roadmap construction. The {\tt IRS} algorithm is an incremental instantation of such a roadmap spanner approach, which only removes nodes. Extensions also remove nodes \citep{dobson2014sparse} resulting in even smaller data structures, which offer
benefits of computational and storage savings. Sparser variants of tree-based planners \citep{li2016asymptotically} also prune nodes that reach similar parts of $\cspace$ with suboptimal paths. 

This line of work has also looked on deciding the number of finite samples for a motion planner in practice by studying the \textit{finite time properties} of \ao methods \citep{dobson2013finite}. This has yielded insights into the trade-offs of computation and solution quality expected from these algorithms.  There have been additional models for studying near-optimality for tree-based approaches \citep{salzman2016asymptotically} and those based on random geometric graphs \citep{solovey2018new, solovey2020critical}.

\textbf{Improving Computational Efficiency:} 
There are many ways to
improve computational efficiency of \ao planners, including through
parallelizing~\citep{bialkowski2011massively} and caching collision
checking \citep{bialkowski2013efficient}. A way to speed up collision checking is to perform it lazily \citep{haghtalab2018provable}. Sampling strategies that focus on regions of $\cfree$ so as to improve existing solutions have been applied to both single-processor planners~\citep{gammell2015batch} and in parallel search processes with shared information~\citep{otte2013c}. Unlike \rrtstar, which uses local rewiring, global cost information propagation~\citep{arslan2013use} allows faster convergence in single-shot and replanning frameworks~\citep{otte2016rrtx}.
Heuristics have also been incorporated into elements of kinodynamic planning~\citep{littlefield2018efficient}.
These optimizations can let \ao
algorithms quickly discover high-quality solutions, while improving on these solutions over time, i.e., they exhibit \textit{anytime} behavior.  Guidance can also arise out of human demonstrations, which guide an underlying \ao search strategy \citep{bowen2016asymptotically}. On real-world systems, computation limits can be sidestepped by using a cloud system \citep{ichnowski2016cloud,bekris2015cloud}.
Hybrid approaches \citep{ChoudhuryGBSS16} have combined optimization strategies to refine the solutions obtained in an \ao framework.

\vspace{0.1in}
\noindent{\bf AO Planners for Extensions of the Basic Problem}

The progress in \ao properties has allowed extending such guarantees
to new domains. In particular, most of the above algorithms are
applicable to kinematic domains given a Euclidean
norm as an optimization objective. Below is a list of effort that extend analysis models to more complex problems.

\textbf{Kinodynamic Planning:} The kinodynamic case deals with motion planning for a robotic system with significant dynamics, i.e., the planning has to consider and account for velocities, accelerations (and other higher order dynamics). The approaches based on random geometric graphs assume the existence of  a \textit{steering function} to guarantee that two nearby configurations can be connected. In a kinodynamic problem, this does not always exist. An \ao approach was proposed for systems with linearized dynamics \citep{webb2013kinodynamic}. Approaches leveraging newer analysis models of Monte Carlo trees \citep{li2016asymptotically, littlefield2018efficient}, and search in the state-cost space \citep{hauser2016asymptotically, kleinbort2019rrt2} have been proposed as \ao algorithmic frameworks in the kinodynamic domain.

\begin{wrapfigure}{r}{0.45\textwidth}
	\vspace{-0.25in}
    \centering
    \includegraphics[width=0.44\textwidth]{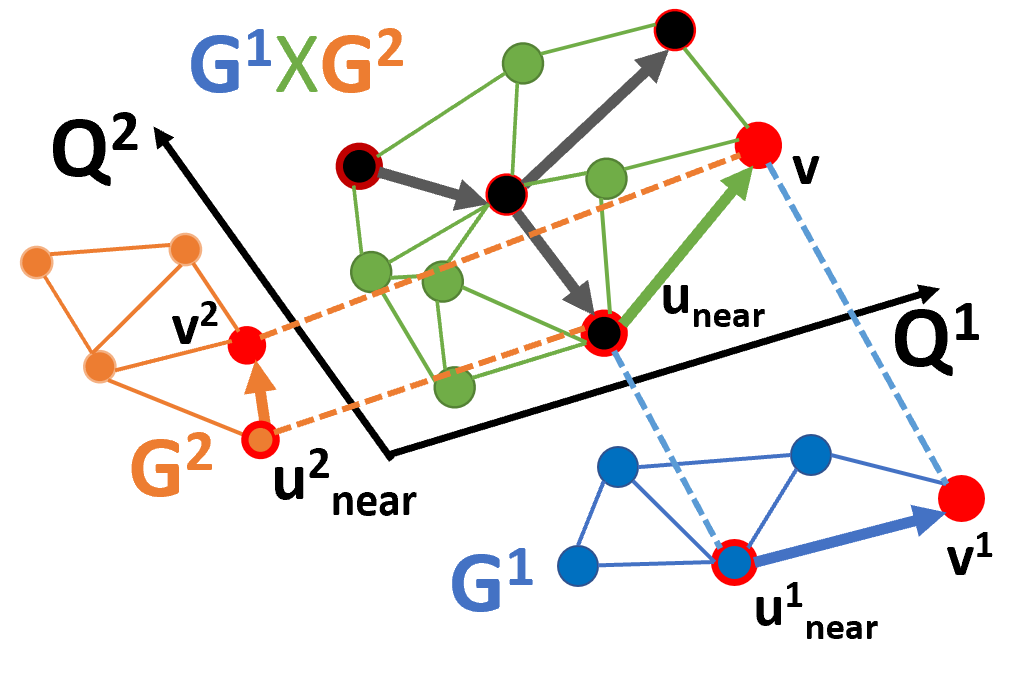}
	\vspace{-0.1in}
    \caption{\small A \textit{tensor-roadmap} $\graph^1\times\graph^2$ constructed from spaces $\cspace^1$ and $\cspace^2$. Black arrows denote the tree growing on the tensor roadmap. An extension to an adjacent vertex is demonstrated with $u_{\mathrm{near}}$ and $\node$.}
	\vspace{-0.3in}
	\label{fig:tensor}
\end{wrapfigure}
\textbf{Multi-robot Motion Planning: } The challenge in this case corresponds to the explosion in dimensionality. Centralized methods
\citep{solovey2015finding} 
have been argued to be \ao \citep{shome2019drrt} under a sampling-based scheme that builds a graph in each robot's $\cfree$ and searches online over an implicit representation of a \textit{tensor-roadmap} of all the robots (shown in Fig. \ref{fig:tensor}). 
The key idea is that the space of $r$ robots can be seen as the Cartesian product of the constituent spaces $\cspace = \cspace^1 \times \cspace^2 \cdots \times \cspace^r$. Accordingly, configurations and nodes can be split into the respective components in the constituent spaces. Fig~\ref{fig:tensor} demonstrates the construction of \ao roadmaps ({\tt ao\_rm}) in each space. The combination of these roadmaps has been shown to be \ao in the entire $\cspace$ and is called the \textit{tensor-roadmap} : $ \graph = \graph^1 \times \graph^2 \cdots \times \graph^r .$ Instead of explicitly creating and storing $\graph$, the idea is to implicitly search it online through a tree $\tree$ that is constructed over the nodes in $\graph$. Avoiding storage and enumeration of the \textit{tensor-roadmap} has important memory and computational benefits.

\textbf{Motion Planning with Constraints: } Many motion planning problems introduce \textit{constraints} that force feasible solutions to lie on lower dimensional manifolds of $ \cspace $. Consider the difference between moving an arm versus moving it \textit{while} keeping a grasped object upright. Such manifolds have 0-volume in the fully dimensional $ \cspace $. This complicates sampling processes. Approaches aim to ensure that samples and connections are found in such domains, while preserving theoretical properties \citep{kingston2018sampling}. Some variants use projection operators to reach the constraint
manifold, while others operate on \textit{tangent-spaces} of the manifold by decomposing local neighborhoods into \textit{atlases} \citep{jaillet2013asymptotically}. A way to look at the problem is to decouple constraint satisfaction from the underlying planner \citep{kingston2017decoupling}. Such constraints also arise in integrated task and motion planning \citep{shomewafrtp,vega2016asymptotically}.

\begin{wrapfigure}{r}{0.5\textwidth}
\vspace{-0.25in}
\begin{tabular}{r|p{0.3\textwidth}}
\textbf{Domain} & \textbf{Algorithms}                                                                                                                                                              \\ \hline
\ao kinematic       & {\prmstar,\rrtstar,\rrg,{\tt RRT}$^\#$, {\tt FMT}$^*$,{\tt BIT}$^*$,{\tt RRT}$^X$}\\ \hline
\ano kinematic      & {{\tt IRS}, {\tt SPARS}, {\tt LBT-RRT}}                       \\ \hline
kinodynamic         & {{\tt SST}$^*$,{\tt AO-}$\mathcal{A}$,{\tt DIRT}, {\tt AO-RRT} } \\ \hline
constraints         & {{\tt atlas-\rrtstar}}                                                                                                      \\ \hline
multi-robot         & {\drrtstar}                                                                                                                             \\ \hline
\end{tabular} 
\vspace{-0.3in}
\end{wrapfigure}

\textbf{Belief-space Planning:} There are efforts in extending properties of sampling-based planners to belief-space planning \citep{agha2012probabilistic, Chaudhari2013}, where instead of planning over individual states, one has
to reason about distributions. Recent work~\citep{littlefield2018importance} has
also demonstrated the considerations and conditions under which such
problems can be solved using \ao algorithms that do not rely on
steering functions.

\section{Examples of Applications}

There has been a push to ensure that methods for specific applications also afford \ao guarantees.  Some example domains (Fig \ref{fig:applications}) where \ao motion planners have been proposed are the following: \textbf{Self-driving cars} \citep{hwan2013optimal}:  \ao planners have been applied to high-speed driving applications on car models with dymamics and fully integrated autonomy systems on shared roadways. \textbf{Space robotics} \citep{littlefield2016integrating}: Space exploration involves deployment in highly unstructured environments. New emerging rovers, such as hybrid soft-rigid mechanisms based on tensegrity, introduce unique planning difficulties, which have been approached with \ao planners. \textbf{Planning for manipulators} \citep{perez2011,schmitt2017,kimmel2018,shome2019multiarm}: Planning for high-dimensional arms requires careful consideration of objects in the scene that need to be reached or avoided. \textbf{Medical robotics} \citep{patil2015}: Robustness and quality of solutions is important in medical tasks and promote \ao considerations. \textbf{Robot design} \citep{baykal2019asymptotically}: The design process for a robot can be seen as a model that, if optimized for a specific application, can significantly help in addressing challenging problems.

\begin{figure}[h]
\vspace{-0.25in}
    \centering
    \includegraphics[height=1in,trim={7cm 0 7cm 0},clip]{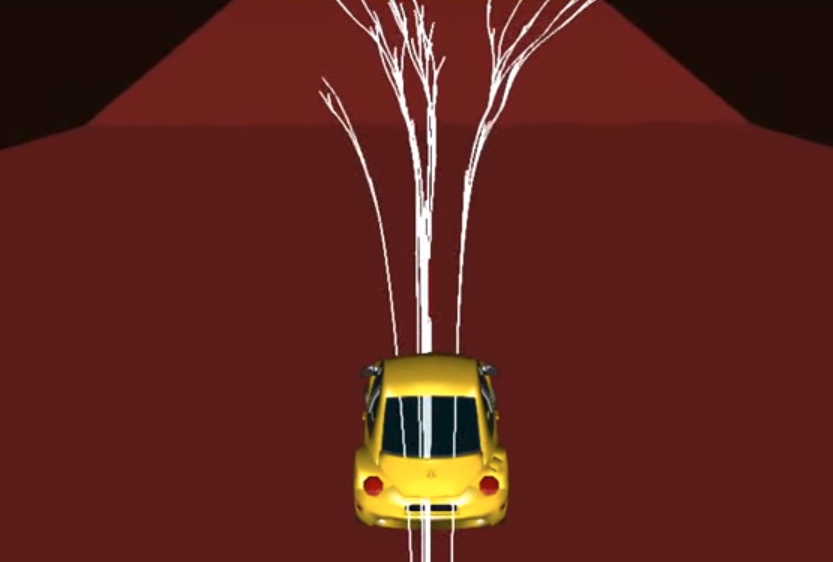}
    \includegraphics[height=1in]{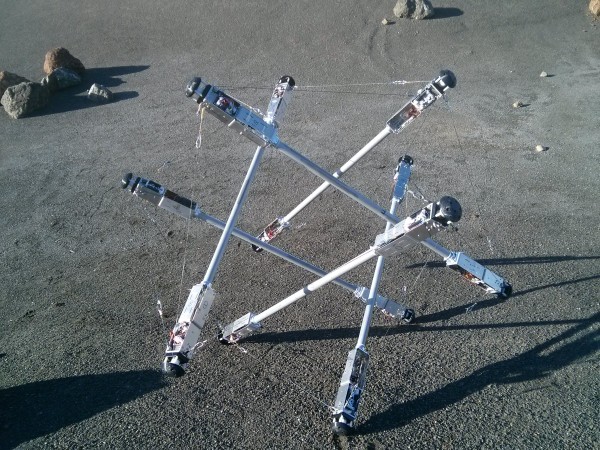}
    \includegraphics[height=1in]{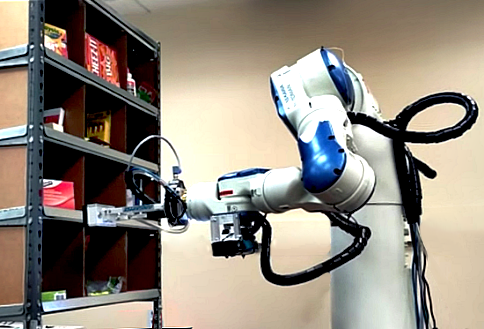}
    \includegraphics[height=1in,trim={4cm 0 4cm 0},clip]{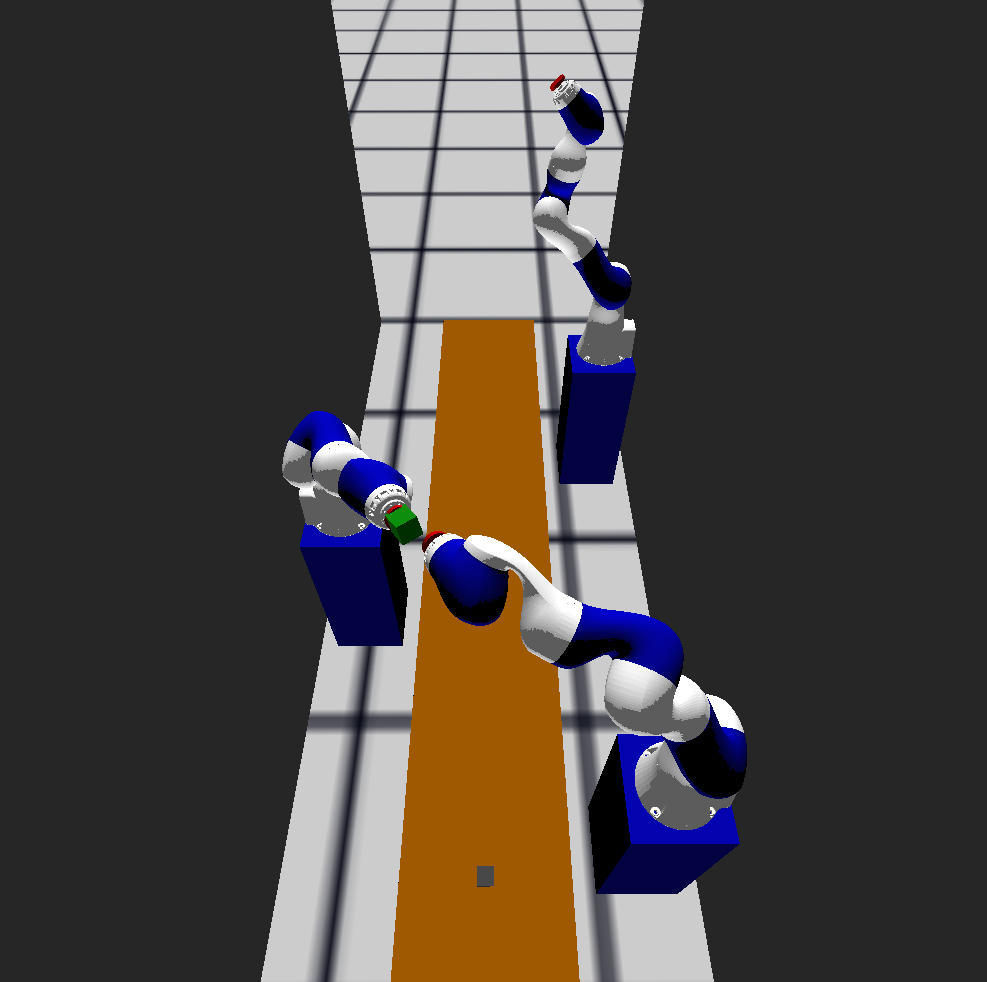}
    \vspace{-.1in}
    \caption{\small (\textit{Left to right}): A car being planned for in simulation \citep{li2016asymptotically}, a \textit{SUPERball} rover for use in space exploration \citep{littlefield2016integrating}, a manipulator planning to grasp objects inside a shelf \citep{kimmel2018}, and a simulation for multi-arm planning~\citep{shome2019multiarm}.}
    \label{fig:applications}
\end{figure}

\section{Future Direction for Research}

Future research can help bridge the gap between theoretical properties and practical performance. For instance, most analyses of sampling-based planners reason for the worst-case. There is the potential for studying the expected behavior of these methods. The inspection of convergence rates as well as efficient data structures can provide insights regarding the practical and predictable deployment of \ao methods.

\textbf{Computing Platforms}: There have been efforts to leverage modern computing hardware, such as parallelization~\citep{bialkowski2011massively} and custom  chipsets~\citep{Murray2016RobotMP} to optimize queries and lead to significant speed-ups. Moreover, future planners will interact with cloud infrastructure and can share information \citep{bekris2015cloud, ichnowski2016cloud}. GPUs, which have led to a revolution in machine learning, can also improve the efficiency of  planners~\citep{ichter2017group}, given efficient parallel primitives~\citep{pan2012gpu}.

\textbf{Learning}:  Learning tools have been integrated with sampling-based planners to speed-up and improve performance of key components \citep{faust2018}. Learning how to perform sampling and identifying latent spaces of complex systems  \citep{ichter2019robot} is an avenue for augmenting motion planners.

\textbf{Planning under Uncertainty}: Further analysis is needed to argue \ao for motion planners in this domain \citep{littlefield2018importance} as well as computational efficiency. 

\textbf{Integrated Task and Motion Planning}: Many domains, such as manipulation, require identifying the sequence of planning problems to be addressed for solving a task. Recent progress has set the foundations for asymptotic optimality in this domain \citep{vega2016asymptotically} and provides opportunities for applying \ao\ planners in problems, such as object rearrangement \citep{shomewafrtp}.

\bibliographystyle{spbasic}

\end{document}